\documentclass[times, review, 10pt]{elsarticle}

\usepackage{amsmath,amsfonts,amssymb}
\usepackage{algorithm}
\usepackage{algpseudocode}
\usepackage{array}
\usepackage{booktabs}
\usepackage{float}
\usepackage[section]{placeins}
\usepackage{graphicx}
\usepackage{textcomp}
\usepackage{url}
\usepackage{verbatim}
\usepackage{xcolor}
\usepackage{soul}
\usepackage[font=footnotesize]{caption}
\usepackage[font=footnotesize]{subfig}
\usepackage{etoolbox}

\journal{Pattern Recognition}

\sethlcolor{yellow}

\AtBeginEnvironment{table}{\normalsize}
\AtBeginEnvironment{tabular}{\normalsize}
\AtBeginEnvironment{algorithm}{\normalsize}


\begin{document}

\begin{frontmatter}

\title{Global Attention-Fused Image Cropping with Attention-Guided and Global-Aligned Crop Evaluator\tnoteref{t1}}

\author[aff1,aff2,aff3]{Haotian Yang}
\ead{yang.haotian@foxmail.com}

\author[aff2,aff3]{Zhile Yang}
\ead{zl.yang@siat.ac.cn}

\author[aff4]{Kin-Man Lam}
\ead{enkmlam@polyu.edu.hk}

\author[aff5]{Patrick Le Callet}
\ead{patrick.lecallet@univ-nantes.fr}

\author[aff1]{Xin Sun\corref{cor1}}
\ead{sunxin1984@ieee.org}
\cortext[cor1]{Corresponding author.}

\affiliation[aff1]{organization={Faculty of Data Science, City University of Macau},
  city={Macao},
  country={China}}

\affiliation[aff2]{organization={Shenzhen University of Advanced Technology},
  city={Shenzhen},
  country={China}}

\affiliation[aff3]{organization={Shenzhen Institute of Advanced Technology, Chinese Academy of Sciences},
  city={Shenzhen},
  country={China}}

\affiliation[aff4]{organization={Department of Electrical and Electronic Engineering, The Hong Kong Polytechnic University},
  city={Hong Kong},
  country={China}}

\affiliation[aff5]{organization={Nantes Université, Ecole Centrale Nantes, CAPACITES SAS, CNRS, LS2N, UMR 6004},
  city={Nantes},
  country={France}}

\begin{abstract}Image cropping aims to improve image aesthetics by preserving important content within an appropriately composed region. However, most existing methods focus primarily on salient regions and therefore have limited sensitivity to the global relationships among the main image components. To address this limitation, we propose Global Attention-Fused Image Cropping (GAFIC), which consists of an Attention-Guided Feature Fusion (AGFF) and a Global-Aligned Crop Evaluator (GACE). AGFF aggregates the importance of local regions to construct a global representation that captures both image structure and local details. GACE aligns candidate crop features with this global representation, enabling crop evaluation to remain sensitive to boundary changes. We further combine three ranking losses across multiple scales to obtain accurate and stable crop scores. Extensive experiments on the GAIC and CPC datasets demonstrate that GAFIC outperforms existing image-cropping methods, particularly in terms of accuracy and stability. Unlike pixel-level retargeting methods such as seam carving, inpainting, and diffusion-based synthesis, GAFIC does not synthesize or modify the retained pixels; instead, it selects an aesthetically preferred crop from the source image, making it suitable for scenarios where pixel integrity and efficient batch processing are important. The source code is available at
\url{https://github.com/AIVRC/GAFIC.git}.
\end{abstract}





\end{frontmatter}

\section{Introduction}

\begin{figure}
  \centering
  \includegraphics[width=\textwidth]{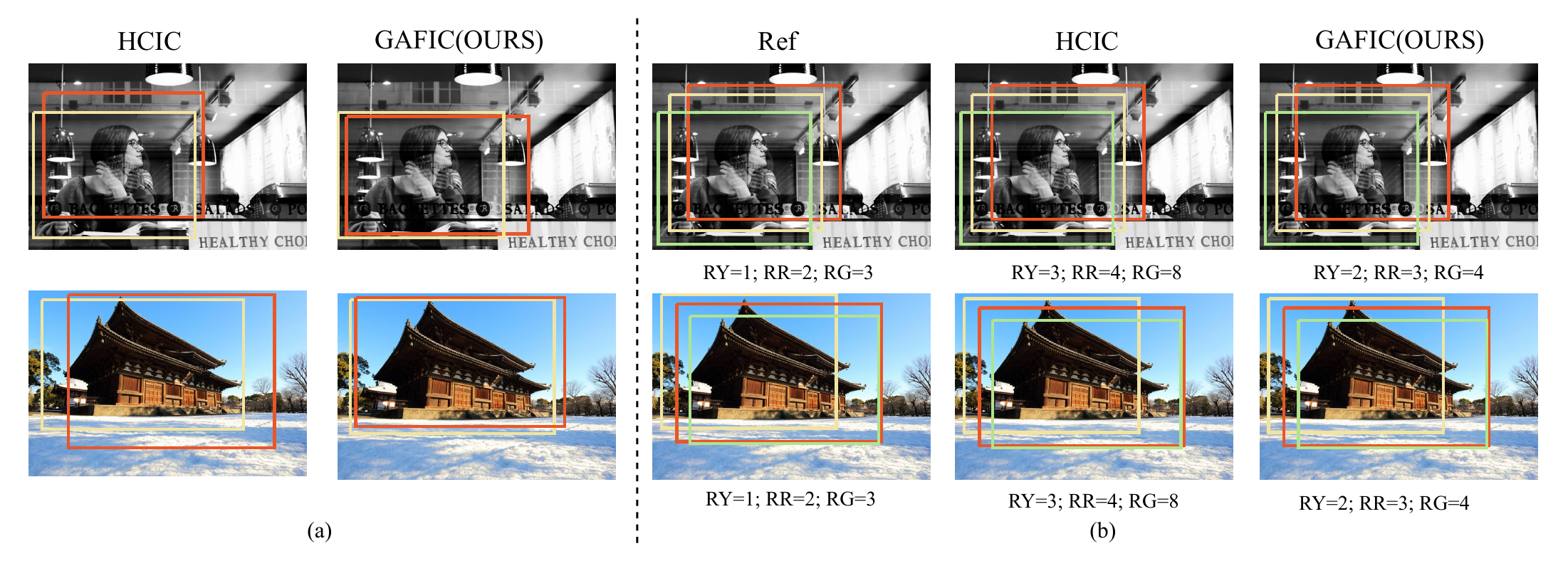}
  \caption{Comparison of representative image-cropping results. (a) HCIC fails to preserve the complete primary content, whereas the proposed method produces a crop that is more consistent with the ground truth. Yellow and red boxes denote the ground-truth and predicted crops, respectively. (b) Red, yellow, and green boxes indicate the first-, second-, and third-ranked manually annotated crops, respectively. The predicted rankings produced by HCIC and the proposed method demonstrate that GAFIC provides greater ranking consistency for
  candidates with similar boundaries.}
  \label{fig:TongDian}
\end{figure}

Image cropping \cite{zeng2020grid} aims to remove redundant information from an image, thereby highlighting important visual components and improving the overall aesthetic quality. The goal is to present an expressive visual effect in a limited space. In recent years, the field of image cropping has made breakthrough progress, with many representative works in candidate evaluation and cropping synthesis, such as Deep Cropping \cite{wang2017deep}, Aesthetic-guided Outward \cite{zhong2021aesthetic}, ClipCrop \cite{zhong2023clipcrop}, Cropper \cite{lee2025cropper}, GAICD \cite{zeng2020grid} and HCIC \cite{zhang2022human}. These methods have greatly improved the aesthetic consistency and compositional rationality of cropping results. Generally, they use attention-dependent features to score candidate windows and select the optimal box with ranking strategies, ensuring strong macroscopic aesthetic performance. However, existing methods typically focus on salient regions to identify the primary component \cite{zeng2020grid, zhang2022human}. They often ignore the importance of various regions within the overall semantics. This can lead to incomplete identification of the primary component or the inclusion of redundant background in cropping boxes \cite{zeng2020grid}. Furthermore, many methods \cite{santella2006gaze,chen2016automatic} focus on overall coverage when selecting cropping boxes, but fail to adequately consider differences between similar boundary regions. Due to insufficient perception of boundary regions, poor ranking stability occurs when cropped images are highly similar. Fig.~\ref{fig:TongDian} illustrates these shortcomings of existing methods, manifested in the generalization of feature representation and insufficient boundary perception.

\textbf{Coarse attention distribution.} Existing methods employ simple attention mechanisms in image feature extraction, failing to reflect the importance of different local regions \cite{zhang2022human}. This results in incomplete principal component recognition or redundant background inclusion in cropped images. As shown in Fig.~\ref{fig:TongDian}(a), HCIC \cite{zhang2022human} fails to accurately capture important content in the cropped image due to coarse attention distribution, e.g., missing details of desktop items or truncating the hand. In contrast, our method clearly preserves the desktop furnishings and gestures while highlighting the object.

\textbf{Poor boundary-regions perception.} Due to insufficient perception of the regions near the cropping boundary, ranking stability deteriorates when cropped images are highly similar. As shown in Fig.~\ref{fig:TongDian}(b), the cropped image rankings (3, 4, and 8) produced by HCIC \cite{zhang2022human} deviate significantly from the manually labeled ideal rankings (1, 2, and 3). In contrast, our method achieves rankings (i.e., 2, 3, and 4), which are closer to the ideal annotation and exhibits strong ranking stability.

To address these challenges, we propose a novel Global Attention-Fused Image Cropping (GAFIC) framework. Specifically, GAFIC introduces Attention-Guided Feature Fusion (AGFF) to model the importance of different image regions and fuse this information with the original feature vector to form a new representation. This representation not only reflects the weights of regions with different importance but also incorporates global information. Subsequently, GAFIC implements the Global-Aligned Crop Evaluator (GACE) to align and fuse features inside and outside the bounding box, as well as the entire image. This enables the model to focus on regions near the cropping boundary while considering overall compositional information when calculating the aesthetic score of the bounding box. Finally, the method synergistically optimizes the best bounding box regression score, the overall bounding box regression score, and the ranking score to improve the aesthetic quality of the cropping result. In summary, our contributions are as follows.

\begin{itemize}
    \item We propose a novel GAFIC framework to address the issues of coarse attention distribution and poor perception of regions near cropping  boundaries during feature representation.
    \item We introduce the Attention-Guided Feature Fusion (AGFF) module, which generates a global feature representing the importance of different regions, thereby mitigating coarse attention distribution.
    \item We propose the novel Global-Aligned Crop Evaluator (GACE), which aligns and fuses cropped information with global feature to enhance perception of border regions.
    \item Extensive comparative experiments on the GAIC and CPC datasets demonstrate the effectiveness of our model compared to several state-of-the-art (SOTA) methods. Our method achieves encouraging results by producing aesthetically pleasing image crops, highlighting the significance of our contribution.
\end{itemize}

The rest of this article is organized as follows: Section \ref{rw} briefly reviews related work. Section \ref{m} presents and discusses the proposed GAFIC framework. Section \ref{e} reports the experimental results and analysis. Section \ref{dc} concludes the article.

\section{Related Work}
\label{rw}
This section briefly summarizes recent progress in related fields, with a particular emphasis on the evolution of image cropping techniques. These approaches range from attention-based and aesthetics-driven methods to data-driven models, ultimately leading to innovative frameworks that integrate generative techniques and advanced feature modeling. Specifically, they can be grouped into the following categories.
\subsection{Retargeting Scope of Image Cropping}
\label{retargeting_scope}
Image retargeting \cite{rubinstein2010comparative} covers a broad spectrum of operations that adapt visual content to target display constraints. Classical retargeting methods, including seam carving \cite{avidan2023seam}, mesh-based warping \cite{wang2008optimized}, and patch-based resizing \cite{lin2012patch}, alter image geometry or remove low-energy paths to satisfy a target aspect ratio. Recent generative methods, such as image inpainting and diffusion-based editing \cite{lugmayr2022repaint,rombach2022high}, can synthesize missing content but introduce new pixels and usually require heavier inference. Image cropping occupies a narrower but practically important position in this spectrum: it does not generate content or deform the retained region, but selects an appropriate subregion from the original image \cite{li2018a2}. This distinction clarifies the application scope of GAFIC. The proposed method is designed for aesthetic crop selection under pixel-preserving constraints, rather than for content synthesis or arbitrary aspect-ratio retargeting.
\subsection{Attention-guided image cropping}
This approach mainly simulates human attention allocation by identifying salient areas in an image, i.e., cropping the most eye-catching part. Early image cropping methods \cite{ fang2014automatic, vikram2012saliency, tong2015salient} utilized saliency detection algorithms to locate the cropping window with the highest attention value, thereby improving image recognizability and visual appeal. For example, Ciocca et al. \cite{ciocca2007self} proposed an adaptive image cropping algorithm that combines a visual attention model with salient area extraction and integrates information such as automatic image classification, face detection and skin color detection to ensure that cropping results align with visual perception habits and user expectations. In order to reduce computational complexity, Chen et al. \cite{chen2016automatic} proposed a simple power function model that automatically selects  the attention value threshold, thereby minimizing computational requirements. Additionally, some methods incorporate user gaze interaction \cite{santella2006gaze} or emotional attention cues \cite{li2019collaborative} to identify important image regions. Although these approaches effectively simulate human visual focus to prioritize key areas, they struggle with low-contrast regions and background details. Moreover, they fail to fully capture areas with potential aesthetic value and cannot adequately measure artistic composition. Consequently, despite the variety of attention-guided image cropping methods, their cropping results often  overlook the overall aesthetic effect.

\subsection{Aesthetics-guided Image Cropping}

This approach evaluates the aesthetic quality of different cropping schemes by learning an aesthetic evaluation model to improve the visual appeal of cropped images. In other words, image quality is assessed comprehensively across multiple dimensions, including composition rules and aspect ratio \cite{zhu2024emotion,liu2025migf, tong2015salient}. Chang et al. \cite{chang2009finding} formulated  image cropping as a search problem, generating candidate cropping boxes and computing structural similarity between candidates  and composition examples. Nishiyama et al. \cite{nishiyama2009sensation} were the first to apply aesthetic evaluation to image cropping. Subsequently, Zhang et al. \cite{zhang2013weakly} characterized aesthetic attributes using handcrafted features or aesthetic composition rules, integrating principles such as attention center, visual salience, image simplicity, facial region overlap, and the rule of thirds compositions \cite{zhang2005auto, liu2010optimizing,yuan2024aesthetic}. Candidate cropping regions are then evaluated using quality measurement methods. Chen et al. \cite{chen2017learning} proposed the VFN image cropping algorithm, which trains an aesthetic evaluation model by comparing different views of the same image. Candidate boxes generated via a sliding window search strategy are scored to determine the final cropping result. Although these methods leverage aesthetic scores effectively, their handcrafted features are constrained by pre-defined aesthetic rules and fail to accurately predict aesthetics in complex images. Furthermore, generating candidate boxes through an aesthetic evaluation network is computationally expensive. Therefore, constructing suitable datasets and optimizing the cropping process remain critical challenges.

\subsection{Data-driven Image Cropping}
Data-driven approaches primarily rely on large amounts of labeled data to train end-to-end deep learning models for image cropping \cite{li2018a2, wang2018deep, lu2019aesthetic}. Without sufficient training samples, these methods usually train a general aesthetic evaluator on image aesthetic databases such as AVA and CUHKPQ for cropping \cite{zeng2020grid}. However, general classifiers cannot reliably evaluate cropped areas within a single image \cite{chen2017learning}. To address this limitation, the CPC \cite{wei2018good} and GAICD \cite{zeng2020grid} datasets were introduced. Wei et al. \cite{wei2018good} proposed the CPC dataset, which contains 12,500 images. For each image, 895 predefined candidate boxes are generated, and 24 cropping boxes are annotated with aesthetic scores. This setup produces over one million pairs with known relative aesthetic rankings, enabling pairwise learning for photo composition. Then using knowledge distillation, a teacher-student aesthetic evaluation network pair, where the teacher network is named VEN and the student network VPN, was trained on CPC. Although the VPN algorithm reduces the time required for image cropping on CPC, redundancy remains due to thousands of candidate boxes. To mitigate this, Zeng et al. \cite{zeng2020grid} proposed GAIC, a grid search-based candidate box generation method that reduces the number of candidate boxes to fewer than 100. The created dataset is called GAICD, which contains 1034 images. For each image, more than 80 candidate boxes labeled with aesthetic scores were generated using the grid search rule. The GAICD method \cite{zeng2020grid} effectively retained the main content while eliminating interference elements by modeling  regions of interest and regions of discard. Zhang  et al. \cite{zhang2022human} used a pairwise learning approach to determine the  scores of different candidate cropping regions. This method introduces a content-preserving module into the human-centered image cropping task, achieving significant improvements in both performance evaluation and preservation  of important content. Additionally, some methods \cite{zhang2020emotion} leverage attention mechanisms or emotional attention to predict key areas in images and combine aesthetic evaluation to optimize and generate cropping results aligned with human aesthetics and composition rules.

Nevertheless, most of the methods extract feature representations from the entire image, overlooking subtle differences between
similar boundary regions. This lack of boundary sensitivity often leads to fluctuations in box rankings when the contents are roughly identical. This paper addresses this issue through global attention fusion. The core innovation is to generate a unified feature that characterizes the global importance of each region, capturing both the overall structure and the expression of local details.
A further distinction among recent data-driven methods lies in the level of representation used for crop evaluation. ClipCrop \cite{zhong2023clipcrop} and Cropper \cite{lee2025cropper} leverage vision-language representations to evaluate image composition at a semantic level. In contrast, GAFIC obtains global awareness through attention-guided feature fusion over a CNN backbone, which keeps the crop-evaluation process aligned with local boundary evidence while avoiding dependence on a large vision-language encoder. This design highlights a quality-efficiency trade-off rather than a direct replacement of CLIP-based semantic evaluation.

\section{Methodology}
\label{m} \noindent

\begin{figure}
    \centering    
    \includegraphics[width=1.0\linewidth]{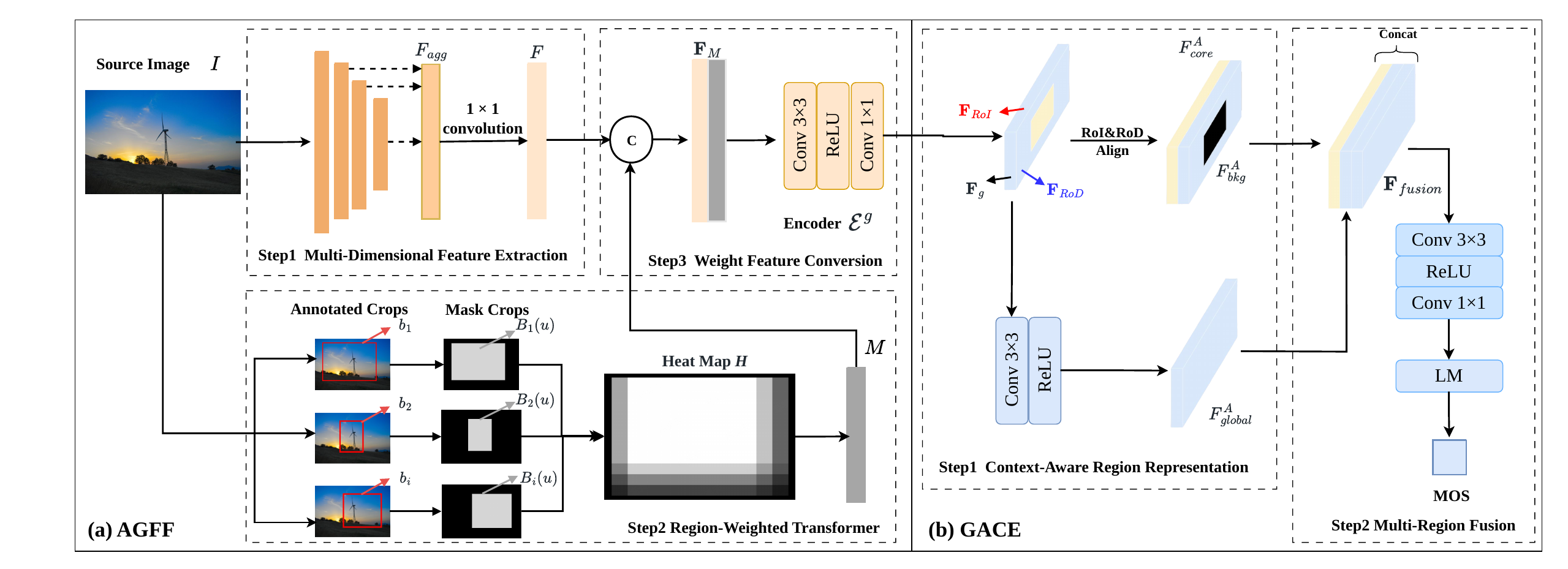}
    \caption{Overall architecture of our proposed GAFIC framework. (a) Attention-Guided Feature Fusion (AGFF) Module; (b) Global-Aligned Crop Evaluator (GACE) Module.}
    \label{fig:JiaGou}
\end{figure}

This section describes the proposed GAFIC framework and explains why it effectively perceives the importance of local regions in aesthetic cropping. Existing methods \cite{wang2017deep, zhang2022human, zhong2023clipcrop} primarily focus on a single salient region, ignoring the importance of different regions. This leads to coarse attention distribution and poor perception of boundary-regions. To address these challenges, GAFIC incorporates two novel components: Attention-Guided Feature Fusion (AGFF), and Global-Aligned Crop Evaluator (GACE). These components enable GAFIC to achieve SOTA performance through a straightforward approach. The overall architecture is shown in  Fig.~\ref{fig:JiaGou}.

\subsection{Attention-Guided Feature Fusion }\label{Attention-Guided Feature Fusion }

It is challenging to accurately extract global features while preserving local details in image aesthetic cropping. Firstly, constructing a multi-dimensional feature representation that captures rich information is difficult. Secondly, there is a lack of systematic solutions for expressing the importance of each region. Finally, maintaining global consistency while focusing on local regions remains an open problem. The proposed AGFF module addresses these limitations in three steps, as illustrated in Fig.~\ref{fig:JiaGou}(a). First, we extract multi-dimensional semantic features from the image to capture comprehensive information. Next, candidate cropping boxes are aggregated into a heat map representing regional importance. Finally, we fuse and regularize the importance heat map with semantic features to obtain a unified representation that combines global semantics and local importance. This fusion strategy improves scoring accuracy in complex scenes by capturing important local information and compensating for the lack of fine-grained feature details.

\textit{1) Multi-Dimensional Feature Extraction }

Given an input image $I \in \mathbb{R}^{H_0 \times W_0 \times 3}$, the backbone network first extracts multi-scale features from $I$. Bilinear downsampling and upsampling are applied to adjust features to different scales, and these features are aggregated to produce a fused feature ${F}_{agg}$. Finally, a $1 \times 1$ convolution $\Phi(\cdot)$ is employed for dimensionality reduction and fusion, yielding the final multi-scale representation, as follows:
\begin{align}
{F} = \Phi({F}_{agg}), \quad{F} \in \mathbb{R}^{H \times W \times C}.
\end{align}

This representation combines local texture information from shallow layers with global semantic information from deep layers, forming the basis for subsequent feature fusion.

\textit{2) Region-Weighted Transformer}

This component converts candidate cropping boxes into a weighted feature map $M$, representing the importance of different image regions. The given set of cropping boxes is denoted as  $\{ b_i \}_{i=1}^{N}$. First, each cropping box $b_i$ has a corresponding binary mask $B_i(u)$. Next, we perform pixel-wise accumulation of all cropping box masks to obtain a heat map $H(u)$, where $u = (x, y)$ represents the spatial coordinates of a pixel. The heat map reflects the importance distribution of regions in the image, providing a weight benchmark for subsequent weighted fusion. Specifically, $B_i(u)=1$ when the cropping box covers the area, otherwise $B_i(u)=0$.
\begin{equation}
H(u)=\sum_{i=1}^{N} B_{i}(u).
\end{equation}

Finally, $H$ is normalized using Min-Max normalization, producing the weighted feature ${M}=\mathrm{Norm}(H)$, where ${M} \in [0, 1]$. 

\textit{3) Weight Feature Conversion }

This step fuses the feature vector $F$ with the weighted feature $M$ to generate a global feature ${F}_{M}$. First, ${M}$ and ${F}$ are concatenated along the channel dimension. The concatenated result is then fed into the encoder $\mathcal{E}^{g}$ to aggregate semantic information from both feature types. Finally, the image fusion feature ${F}_{g}$ is obtained as follows:
\begin{align}
&F_M = [F, M], \\
&F_g = \mathrm{Conv}_{1\times1}\!\big(
        \mathrm{ReLU}\mathrm({Conv}_{3\times3}(F_M))
      \big).
\end{align}
where $[F,M]$ denotes the concatenation operation of $F$ and $M$, $F_{g}$ denotes the image fusion feature. ${Conv}_{3\times3}(\cdot)$ and
${Conv}_{1\times1}(\cdot)$ represents convolution operations with kernel sizes of $3\times3$ and $1\times1$, respectively, and
${ReLU}(\cdot)$ is the activation function.

\subsection{Global-Aligned Crop Evaluator}\label{Global-Aligned Crop Evaluator}
One major challenge in image aesthetic cropping is ensuring that candidate cropping boxes capture global important information while remaining sensitive to areas near the boundaries.
Therefore, it is critical to simultaneously consider the whole image, the content within the cropping box, and the background outside the box. Moreover, a systematic approach is needed to extract global information that fully reflects the overall compositional tone of the image. Finally, the method should guarantee that each cropping box perceives the main content by integrating the global information.

To overcome these challenges, we propose the GACE module, as shown in Fig. ~\ref{fig:JiaGou}(b). The core process of GACE consists of two steps. First, we extract features from the inner and outer regions of each candidate cropping box. By aligning and merging these two regions, we obtain overall contextual information. Simultaneously, global information is extracted from the fused feature $F_g$ using a global sampler. Next, we fuse the overall contextual information with global features to obtain the multi-region fusion feature  $F_{fusion}$. This feature is then passed through two fully connected layers to predict the aesthetic score of the cropping box.

\textit{1)  Context-Aware Region Representation }
As shown in Fig. \ref{fig:JiaGou}(b), ${F}_{g}$ represents the image fusion feature output by AGFF, while ${F}_{RoI}$ and ${F}_{RoD}$ represent the features within RoI and RoD, respectively. First, we  uniformly extract key internal feature $F_{\text{core}}^A$ and external feature $F_{\text{bkg}}^A$ for each cropping box \cite{zhang2022human}. Next, these two features are concatenated, as follows:
\begin{align}
F^A_{\alpha} = [F{\substack{A\\core}},F{\substack{A\\bkg}} ].
\end{align}
where $F_{\text{core}}^A$ denotes the sampling result within the cropping box, and $F_{\text{bkg}}^A$ denotes the sampling of the remaining area outside the cropping box. To extract global features from $F_g$, we instantiate the transformation function $\Phi(\cdot)$ as a $3 \times 3$ convolutional layer with $C_0$ output channels, followed by the ReLU activation function. The process is formulated as follows:
\begin{equation}
F_{\text{global}}^A = \Phi(F_g).
\end{equation}

\textit{2) Multi-Region Fusion }

This module aims to achieve effective feature fusion and final score prediction. First, it concatenates the aligned feature $F_{\alpha}^A$ and the global feature $F_{global}^A$, to generate a unified fusion feature $F_{fusion}$.
\begin{equation}
F_{fusion} = [F_{\alpha}^A, F_{global}^A].
\end{equation}

Next, the fused feature map is further refined by a lightweight convolutional block and then fed into a linear mapping module for the final Mean Opinion Score (MOS) prediction:
\begin{align}
&\tilde{F} = \mathrm{Conv}_{1\times1}\!\big(
        \mathrm{ReLU}(\mathrm{Conv}_{3\times3}(F_{\mathrm{fusion}}))
    \big), \\
&S_{\mathrm{aes},ij} = f_{\mathrm{LM}}(\tilde{F}).
\end{align}
where ${Conv}_{3\times3}(\cdot)$ and
${Conv}_{1\times1}(\cdot)$  denote convolution operations with kernel sizes of  $3\times3$ and $1\times1$, respectively. ${ReLU}(\cdot)$ is the activation function. $\tilde{F}$ is the refined
feature, and $f_{LM}(\cdot)$ represents the linear
mapping (LM) regression head that pools the feature and produces the aesthetic prediction score $S_{\mathrm{aes},ij}$.

\subsection{Optimization Objective}\label{Optimization  Objective}

Finally, we need to design an optimization objective that guarantees both accurate score predictions and stable ranking predictions. Firstly, the accuracy of the optimal cropping box prediction is an intuitive metric for evaluating the model. Therefore, incorporating this metric into the optimization objective is essential. Secondly, robustness is a key criterion for model evaluation, where the average accuracy of scores and ranking stability are crucial factors reflecting model robustness. Lastly, a unified framework is required to encompass both the predictive performance of the optimal candidate box and the overall robustness of the model.

To address these challenges, we propose a multi-dimensional loss optimization strategy. First, we introduce the optimal score loss function $\mathcal{L}_{BestReg}$, which measures spatial overlap. The spatial overlap rate is defined as the intersection area between the candidate box ${b}_{best}$ (corresponding to the optimal score) and the annotated box ${b}_{label}$, divided by the union area of these two boxes. Next, we introduce an overall score loss ($\mathcal{L}_{Reg}$) and a ranking loss ($\mathcal{L}_{Rank}$). $\mathcal{L}_{Reg}$ represents the difference between the average score and the annotated score across all candidate cropping boxes, while $L_{Rank}$ represents the difference between the predicted ranking and the annotated ranking among candidate cropping boxes. These two loss functions enable the model to optimize for both overall accuracy and ranking stability, thereby improving robustness. Finally, we design a weighted overall loss function $\mathcal{L}_{opt}$ to address the joint optimization problem. This function uses weights to balance the three objectives, ensuring that they work synergistically.

\textit{1)  Best Regression Loss }

The best regression loss ($\mathcal{L}_{BestReg}$) enforces consistency between the optimal prediction box and the annotated optimal box, as follows:
\begin{equation}
\mathcal{L}_{BestReg} = 1 - \frac{S({b}_{best} \cap {b}_{label})}{S({b}_{best} \cup {b}_{label})}
\end{equation}
where ${b}_{best}$ is the cropping box with the highest prediction score, ${s}_{aes,ij}$, ${b}_{label}$ is the manually annotated optimal box, and $\boldsymbol{S}(\cdot)$ represents the area of the region. 

\textit{2)  Regression Loss}

The  regression loss ($\mathcal{L}_{Reg}$) minimizes the discrepancy between the predicted ratings and the manually annotated ratings, as follows:
\begin{equation}
\mathcal{L}_{\text{Reg}} =
\begin{cases}
  \frac{1}{2} (S_{\text{aes},ij} - g_{ij})^2, & |S_{\text{aes},ij} - g_{ij}| \le \delta. \\
  \delta |S_{\text{aes},ij} - g_{ij}| - \frac{1}{2} \delta^2, & \text{otherwise}.
\end{cases}
\label{eq:reg损失函数}
\end{equation}
where ${s}_{aes,ij}$ is the model's predicted score of the $j$-th crop for image $i$, ${g}_{ij}$ is the annotation score, and $\delta = 1$  is the error switching point. This ensures that predicted scores remain consistent with annotated scores.

\textit{3)  Ranking Loss }

The ranking loss ($\mathcal{L}_{Rank}$) ensures that high-quality proposals receive higher scores than low-quality ones \cite{chen2017learning}, as follows:
\begin{equation}
\mathcal{L}_{\text{Rank}} = \sum_{i} \sum_{(j,k): y_{jk}=1} \max(0, \Delta - (S_{\text{aes},ij} - S_{\text{aes},ik})).
\end{equation}
where $y_{jk} = 1$ means that, for the same image $i$, cropping \textit{box} $j$ is confirmed to be better than cropping \textit{box} $k$ by manual annotation, and $\Delta$ =1 is the minimum significant difference threshold.

\textit{4) Total Loss }

Finally, the total loss function ($\mathcal{L}_{opt}$) achieves joint optimization of scoring accuracy, ranking consistency, and optimal result matching, as follows:
\begin{equation}
\mathcal{L}_{opt} = \lambda \cdot\mathcal{L}_{Best Reg} + \mathcal{L}_{Reg} +  \mathcal{L}_{Rank}.
\label{eq:opt损失函数}
\end{equation}
where $\lambda$ is the trade-off parameter.

\section{Experiments}
\label{e}

\begin{table}[htbp]
\centering
  \normalsize
\begin{tabular}{lccccc}
\toprule
Method & Training & IoU$\uparrow$ & Disp$\downarrow$ & FLOPs & Parameters \\
\midrule
VEN \cite{wei2018good} & CPC & 0.837 & 0.041 & 15.39G & 40.93M \\
ASM-Net \cite{tu2020image} & CPC & 0.849 & 0.039 & 64.36G & 14.95M \\
LVRN \cite{wang2018deep} & CPC & 0.843 & - & 15.39G & 40.93M \\
GAIC \cite{zeng2020grid} & GAICD & 0.834 & 0.041 & 20.07G & 16.31M \\
CGS \cite{li2020composing} & GAICD & 0.836 & 0.039 & 20.08G & \ 21.25M \\
CAC-Net \cite{hong2021composing} & FCDB & \textbf{0.854} & 0.033 & 16.26G & 18.93M \\
HCIC \cite{zhang2022human} & CPC & 0.850 & 0.034 & 20.25G & 19.47M \\
GAFIC (Ours) & CPC & 0.853 & \textbf{0.031} & 19.87G & 18.21M \\
\bottomrule
\end{tabular}
\caption{Computational complexity comparison on the whole FLMS dataset.}
\label{tab:complexity}
\end{table}



\subsection{Implementation Details}

We implemented the model using PyTorch and optimized it with the ADAM optimizer. The model was trained for 80 epochs with a fixed learning rate of 0.0001. Following Zeng et al. \cite{zeng2020grid}, we resized the short side of all input images to 256 pixels. A pre-trained VGG16 network \cite{simonyan2014very} was used as the backbone to obtain the base feature $F_{agg}$. Our multidimensional feature extraction used $1\times1$ convolutions to generate a low-dimensional feature map $F$. We then fused $F$ with a heatmap $M$ and generated the main 256-dimensional feature map $F_{M}$ using an encoder. RoIAlign \cite{he2017mask} and RoDAlign \cite{zeng2020grid} were applied to extract the core ($F_{\text{core}}^A$) and background ($F_{\text{bkg}}^A$) regions of candidate cropping boxes as 256-dimensional features \cite{zhang2022human}. These features were fused with aligned global features, and the resulting $9 \times 9 \times 640$ tensors were passed to a two-layer scoring network.
Table~\ref{tab:complexity} reports the computational complexity of the compared image-cropping methods. We train GAFIC on the CPC training set and evaluate IoU and Disp on the FLMS dataset for quantifying computational performance. FLOPs are calculated using the input resolution adopted by each method. Under this setting, GAFIC requires 19.87G FLOPs and 18.21M parameters. GAFIC operates within the same order of magnitude as other CNN-based methods and achieves competitive computational complexity.

\subsection{Comparison to State-of-the-Art Methods}

To demonstrate the effectiveness of our method, we conducted a comprehensive evaluation. We compared our method, GAFIC, with state-of-the-art image cropping methods, including VFN \cite{chen2017quantitative}, VEN \cite{wei2018good}, ASM-Net \cite{tu2020image}, GAIC \cite{zeng2020grid}, CGS \cite{li2020composing}, HCIC \cite{zhang2022human}, SAC-Net \cite{yang2023focusing} and Cropper \cite{lee2025cropper}. For quantitative evaluation, we used multiple metrics, including the average Spearman Rank Correlation Coefficient ($\overline{SRCC}$) \cite{zeng2020grid}, average top-N Accuracy ($\overline{Acc_N}$) \cite{zeng2020grid}, Intersection-over-Union (IoU) \cite{chen2017learning} and Boundary-displacement-error (Disp) \cite{mayer2016large}. 
$\overline{SRCC}$ evaluates the linear correlation between the mean opinion score (MOS) of the predicted cropping boxes and the annotated crop box rankings. $\overline{Acc_N}$ measures the average accuracy of whether the top $N$ predicted cropping boxes are included among the top $N$ annotated cropping boxes. Specifically, $\overline{Acc_5}$ and $\overline{Acc_{10}}$ assess the ability to return the top 5 and top 10 cropping boxes, respectively. IoU evaluates the overlap between predicted cropping boxes and annotated results, while Disp calculates the average distance between the annotated boundary coordinates and the predicted cropping box boundary coordinates. In addition, we conducted detailed qualitative evaluations. To ensure fairness, we used publicly available source code for all comparison methods without fine-tuning, ensuring identical experimental conditions and eliminating potential bias. 
These metrics evaluate geometric precision (IoU and Disp) and ranking consistency ($\overline{SRCC}$ and $\overline{Acc_N}$), but they do not directly measure the semantic coherence of the retained region or the perceptual quality of the attention distribution. This is a common limitation of current image-cropping benchmarks. Complementary evaluations using semantic-similarity or attention-based measures could further strengthen the analysis of attention distribution and are left for future work. In addition, the current protocol aggregates all test images without stratifying them by annotation agreement. For images with multiple plausible crops, a single optimal ranking may be inherently ambiguous, so future work may analyze performance under different levels of crop-quality consensus.

\subsubsection{Quantitative Evaluation}\quad

To fully validate the effectiveness of our method, we performed quantitative comparisons with SOTA approaches across multiple public datasets. To validate the accuracy of our proposed image cropping method, we tested it on the FCDB \cite{chen2017quantitative} dataset. Since ranking stability depends on multiple candidate cropping boxes per image, and the FCDB dataset only provides the best cropping result for each image, we additionally evaluated ranking on the GAICD dataset. Furthermore, to assess the generalizability of our method, we compared it with a human-based image cropping method \cite{zhang2022human} on the CPC dataset. 

Specifically, the FCDB dataset has 3,413 images, with 3,065 for training and 348 for testing. "Training-free" simply denotes that the model is not subjected to any dataset-based training, since Cropper \cite{lee2025cropper} is developed upon a large-scale vision-language model. As demonstrated in Table~\ref{tab:SACD}, our method achieved an IoU value of 0.762, which is the best among all comparison algorithms. This strongly demonstrates that the best cropping box returned by our method has high overlap with the ground truth. Furthermore, our method also achieved a Disp value of 0.051, again the best among all comparison algorithms. This further verifies that we achieve a small pixel-level offset of the cropping box boundaries.

\begin{table}
\centering
  \normalsize
\setlength{\tabcolsep}{5pt} 
   \begin{tabular}{lcccc}
        \toprule
        Method & Source & Training & IoU↑ & Disp↓ \\
        \midrule
        A2RL \cite{li2018a2} & CVPR'18 & FCDB & 0.695 & 0.073 \\
        VPN \cite{wei2018good} & CVPR'18 & FCDB & 0.716 & 0.068 \\
        GAIC \cite{zeng2020grid} & TPAMI'20 & FCDB & 0.673 & 0.064 \\
        Mars \cite{li2020learning} & CVPR’20 & FCDB & 0.735 & 0.062 \\
        Cropper \cite{lee2025cropper} & CVPR'25 & Training-free & 0.756 & 0.053 \\
        \midrule
        Ours & -& FCDB & \textbf{0.762} & \textbf{0.051} \\
        \bottomrule
    \end{tabular}
\caption{Comparison results on the FCDB \cite{chen2017quantitative} dataset.}
\label{tab:SACD}
\end{table}

The GAICD dataset contains 3,336 images, with 2,636 for training, 200 for validation, and 500 for testing. Table~\ref{tab:wendingxing} illustrates that our GAFIC method achieved an $\overline{SRCC}$ of 0.906, with $\overline{Acc_5}$ and $\overline{Acc_{10}}$ scores of 84.5 and 96.6, respectively. These results demonstrate the exceptional ranking stability of our approach, even when multiple cropping boxes share substantial content similarity.
\begin{table}
\centering
  \normalsize
\setlength{\tabcolsep}{2pt} 
    \begin{tabular}{lccccc}
        \toprule
        Method & Source  & Training & $\overline{SRCC}$↑ & $\overline{Acc_5}$↑ & $\overline{Acc_{10}}$↑ \\
        \midrule
        VFN \cite{chen2017quantitative} & MM'17  & GAICD & 0.485 & 26.4 & 40.1  \\
        VEN \cite{wei2018good} & CVPR'18  & GAICD & 0.616 & 35.5 & 48.6 \\
        GAIC \cite{zeng2020grid} & TPAMI'20  & GAICD & 0.849 & 63.1 & 81.6  \\
        CGS \cite{li2020composing} & CVPR'20  & GAICD & 0.795 & 59.7 & 77.8 \\
        TransView \cite{ pan2021transview}  & ICCV'21  & GAICD & 0.857 & 63.9 & 82.4 \\
        Chao et al. \cite{wang2023image} & CVPR'23 & GAICD & 0.872 & 64.8 & 83.3  \\
        Cropper \cite{lee2025cropper} & CVPR'25 &  Training-free & 0.904 & 84.3 & 96.5  \\
        \midrule
        Ours & - & GAICD & \textbf{0.906} & \textbf{84.5} &  \textbf{96.6}\\
        \bottomrule
    \end{tabular}
\caption{Comparison results on the GAICD \cite{zeng2020grid} dataset. }
\label{tab:wendingxing}
\end{table}

We also trained our model on 1,154 human-centric images extracted from the CPC dataset (total 10,797 images). For testing, we selected 176 and 39 human-centered images from the FCDB and FLMS \cite{fang2014automatic} datasets, respectively. As presented in Table~\ref{tab:CPC_jingzhundu}, our algorithm achieved IoU and Disp metrics of 0.7689 and 0.0512, respectively, outperforming existing human-centric cropping methods. This substantiates the leading-edge performance of our approach in specialized image cropping tasks.

\begin{table}
\centering
  \normalsize
\setlength{\tabcolsep}{8pt}
   \begin{tabular}{lcccc}
        \toprule
        Method & Source & Training & IoU↑ & Disp↓ \\
        \midrule
        VFN \cite{chen2017quantitative} & MM'17 & CPC & 0.6509 & 0.0876 \\
        VEN \cite{wei2018good} & CVPR'18 &  CPC & 0.6670 & 0.0837 \\
        ASM-Net \cite{tu2020image} & AAAI'20 &  CPC & 0.7084 & 0.0755 \\
        GAIC \cite{zeng2020grid} & TPAMI'20 &  CPC & 0.7260 & 0.0708 \\
        CGS \cite{li2020composing} & CVPR'20 &  CPC & 0.7331 & 0.0689 \\
        HCIC \cite{zhang2022human} & ECCV'22 &  CPC & 0.7469 & 0.0648 \\
        \midrule
        Ours & -&  CPC & \textbf{0.7689} & \textbf{0.0512} \\
        \bottomrule
    \end{tabular}

\caption{Comparison results on the FCDB \cite{chen2017quantitative} and FLMS \cite{fang2014automatic} datasets.  }
\label{tab:CPC_jingzhundu}
\end{table}

\subsubsection{Qualitative Evaluation}\quad

\begin{figure}
    \centering
    \includegraphics[width=1\columnwidth]{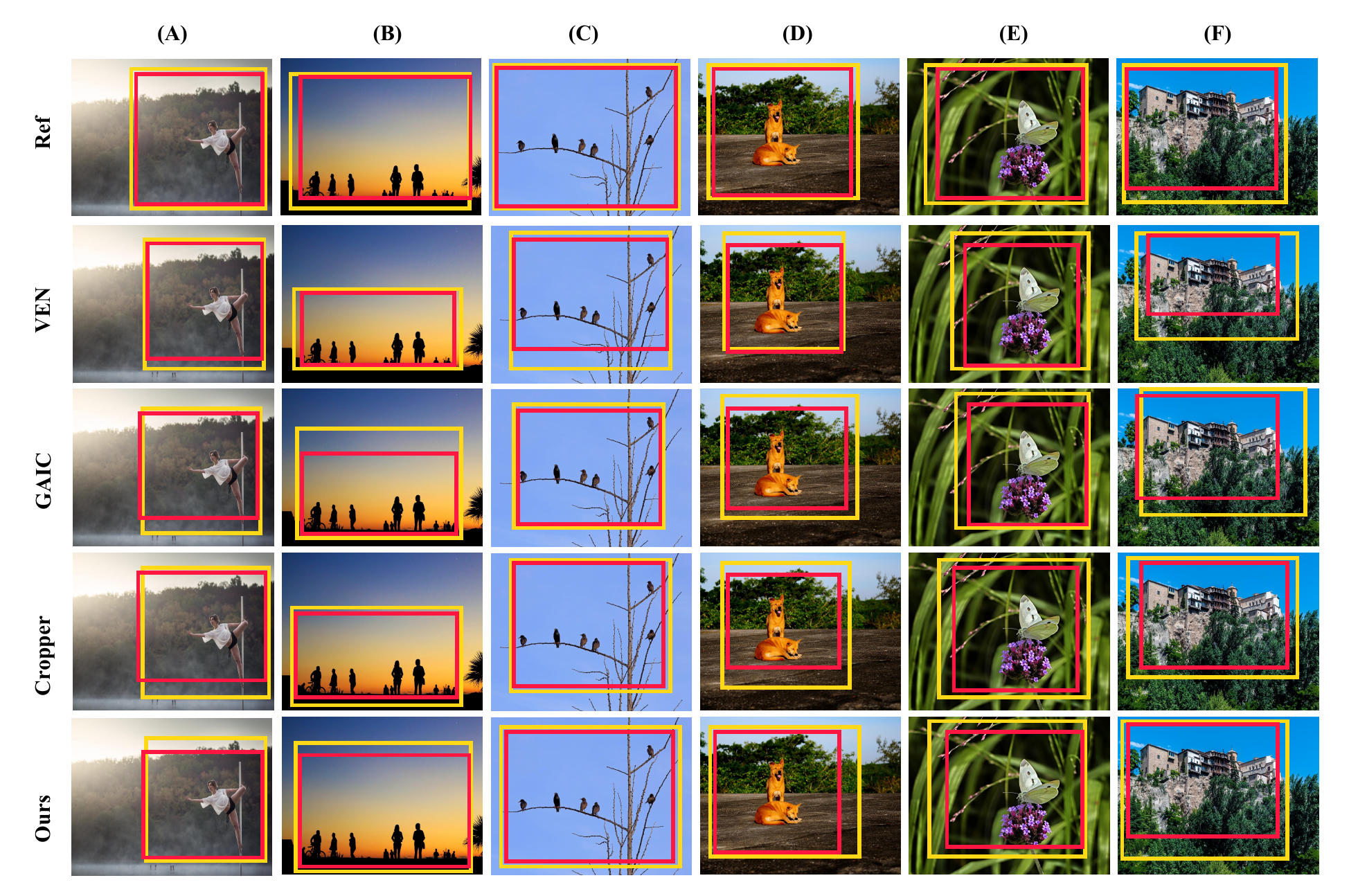}
    \caption{In a qualitative comparison of the top-1 cropping boxes returned, we present the existing state-of-the-art methods side by side with our method, where the yellow cropping boxes represent the ground truth and the red cropping boxes represent the prediction results.}
    \label{fig:mianliao}
\end{figure}

\begin{figure}
    \centering
    \includegraphics[width=1\columnwidth]{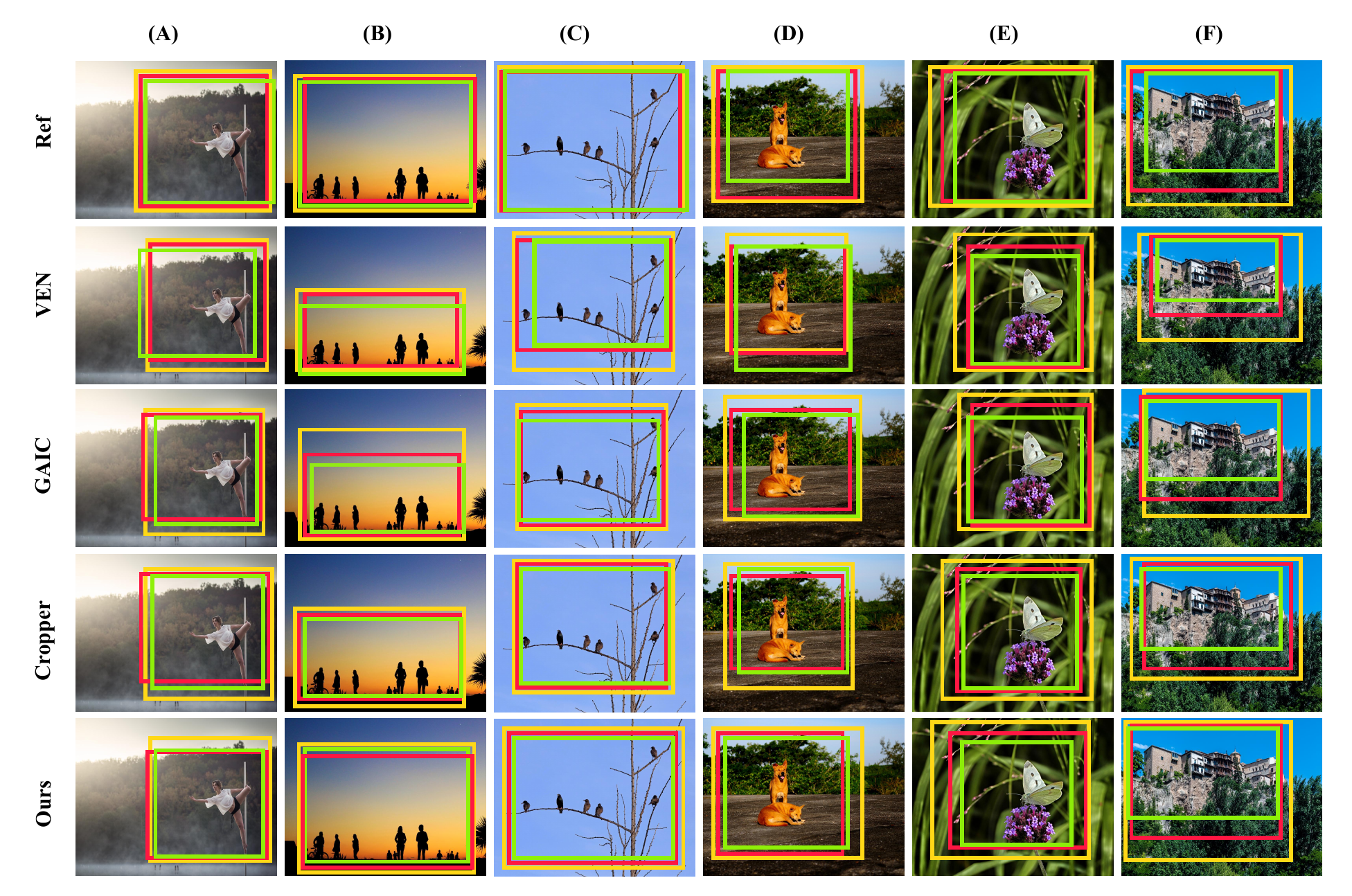}
    \caption{This figure shows the top three predictions returned by each method. The red box represents the best crop planting area, the yellow box represents the second best crop planting area, and the green box represents the third best crop planting area.}
    \label{fig:wezi}
\end{figure}

To further verify the effectiveness of the proposed framework, we conducted a comparative analysis from the perspective of visual aesthetic feedback. Our evaluation focused on two critical aspects: the model’s ability to express the importance of different local regions and its capacity to perceive boundary areas of cropping boxes. The first aspect is examined through coarse-grained attention distribution maps, which reveal how the model prioritizes various regions within an image. The second aspect addresses the stability of candidate cropping box rankings when their contents exhibit substantial similarity. Figs.~\ref{fig:mianliao}--~\ref{fig:wezi} show the visual comparison results of VEN \cite{wei2018good}, GAIC \cite{zeng2020grid} and Cropper \cite{lee2025cropper} under identical conditions.

In order to demonstrate the universality of image cropping, we selected images from different categories, including landscapes, portraits, animals, plants, distant views, close-ups, with various styles and angles. We also provide examples to analyze details of different focal points.

\textbf{In terms of coarse attention distribution.} 
Fig. \ref{fig:mianliao} shows the optimal cropping results for each model. The yellow boxes indicate the annotated ground truth, and the red boxes represent predicted cropping results. We can clearly observe that, for images with clear and distinct objects in columns (A), (D), and (E), our method consistently delivers results that closely match the ground truth. For example, in image (A), VEN \cite{wei2018good} truncates the foot movement when the person is centered, whereas our method preserves the details of the foot movement while highlighting the object. A similar pattern is observed for images with blurred objects in columns (B), (C), and (F). For example, GAIC \cite{zeng2020grid} truncates the lower edge of the purple flower in the close-up photo (E), while our method retains the entire object. Overall, our method demonstrates accurate and effective cropping performance.

\textbf{In terms of boundary area perception.} 
Fig.~\ref{fig:wezi} shows the top three predictions for each model. The first row shows the annotated boxes of the reference image. We can clearly observe that for images in columns (A), (D), and (E) with obvious objects, our method’s top three predictions are closer to the ground truth than those of other methods. For example, in image (D), the proportions and areas of the candidate boxes predicted by VEN \cite{wei2018good} deviate significantly from reality. The same pattern is evident in images without obvious objects in columns (B), (C), and (F). For example, in image (C), the candidate box predicted by GAIC \cite{zeng2020grid} differs significantly from the annotated box, whereas the top three candidate boxes predicted by our method are much closer to the ground truth.

In general, GAFIC outperforms existing methods in both quantitative and qualitative evaluations. This superiority stems from the model’s enhanced ability to express coarse attention distribution across different regions and its improved perception of  boundary areas. These advantages make GAFIC more suitable for general aesthetic image cropping tasks.
\begin{figure}
    \centering
    \includegraphics[width=1\columnwidth]{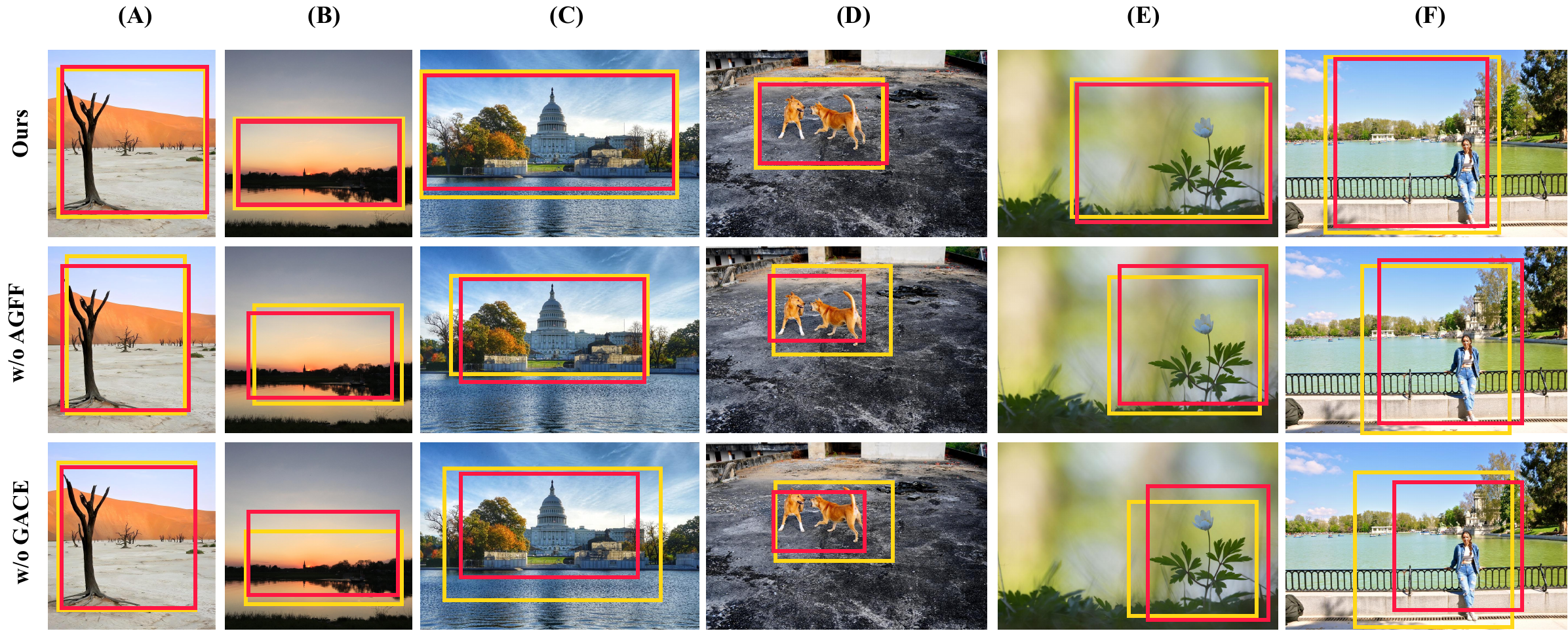}
    \caption{Qualitative comparison of the top-1 cropping boxes returned by the model under different ablation settings. The yellow box represents the ground truth, while the red cropping box represents the prediction result.}
    \label{fig:ablation}
\end{figure}

\subsection{Ablation Study}

We conducted an ablation experiment to evaluate the contributions of the core components, AGFF and GACE, in our GAFIC framework. Our goal is to gain a deeper understanding of the impact of each module on two key performance dimensions: coarse-grained attention distribution across different regions and perception capability in boundary regions.

\subsubsection{Quantitative Evaluation}\quad

We first conducted quantitative ablation experiments to evaluate the specific impact of each component on cropping quality. To ensure fairness, we selected 2,636 images from the GAICD dataset for training, 200 for validation, and 500 for testing to verify ranking stability. Additionally, to simulate user cropping behavior, we selected 3,065 images from the FCDB dataset for training and 348 for testing. The results are shown in Table~\ref{tab:xiaorong}. In the baseline setting (Row 1), only features extracted from the base feature map were used to predict the score for each cropping region. Row 2 evaluated the proposed attention-based heatmap feature fusion module, which improved the $\overline{Acc_5}$  metric by 3.4 and the IoU metric  by 0.058. Row 3 evaluated the proposed global-aligned crop evaluator, which improved $\overline{Acc_5}$ by 5.5 and IoU by 0.102. These results demonstrate that our proposed components effectively improve the accuracy of feature representation, aesthetic scoring, and ranking stability. Furthermore, Row 4 shows that combining both modules, i.e., our method, achieves substantially better performance than adding each module individually, further validating the effectiveness of our method.

\begin{table}[ht]
\centering
  \normalsize
\setlength{\tabcolsep}{3pt} 
    \begin{tabular} {lccccccc}
        \toprule
       Num & AGFF & GACE  & $\overline{SRCC}$↑ & $\overline{Acc_5}$↑ &$\overline{Acc_{10}}$↑ & IoU↑ & Disp↓ \\
        \midrule
        1 &  &  &   0.813  & 78.1  & 90.9 & 0.573 &  0.0628\\
        2 & \checkmark &  &    0.865 & 81.5  & 92.3 & 0.631 & 0.0558 \\
        3 &  & \checkmark   & 0.847  & 83.6  &  94.6& 0.675 & 0.0575\\
        4 & \checkmark & \checkmark & 0.906 & 84.5  &  96.6 &  0.762 & 0.051 \\
        \bottomrule
    \end{tabular}
\caption{Quantitative evaluation results of our method with different module configurations on the GAICD \cite{zeng2020grid} and FCDB \cite{chen2017quantitative} datasets.}
\label{tab:xiaorong}
\end{table}

\subsubsection{Qualitative Evaluation}\quad

\begin{figure}
    \centering
    \includegraphics[width=1.0\columnwidth]{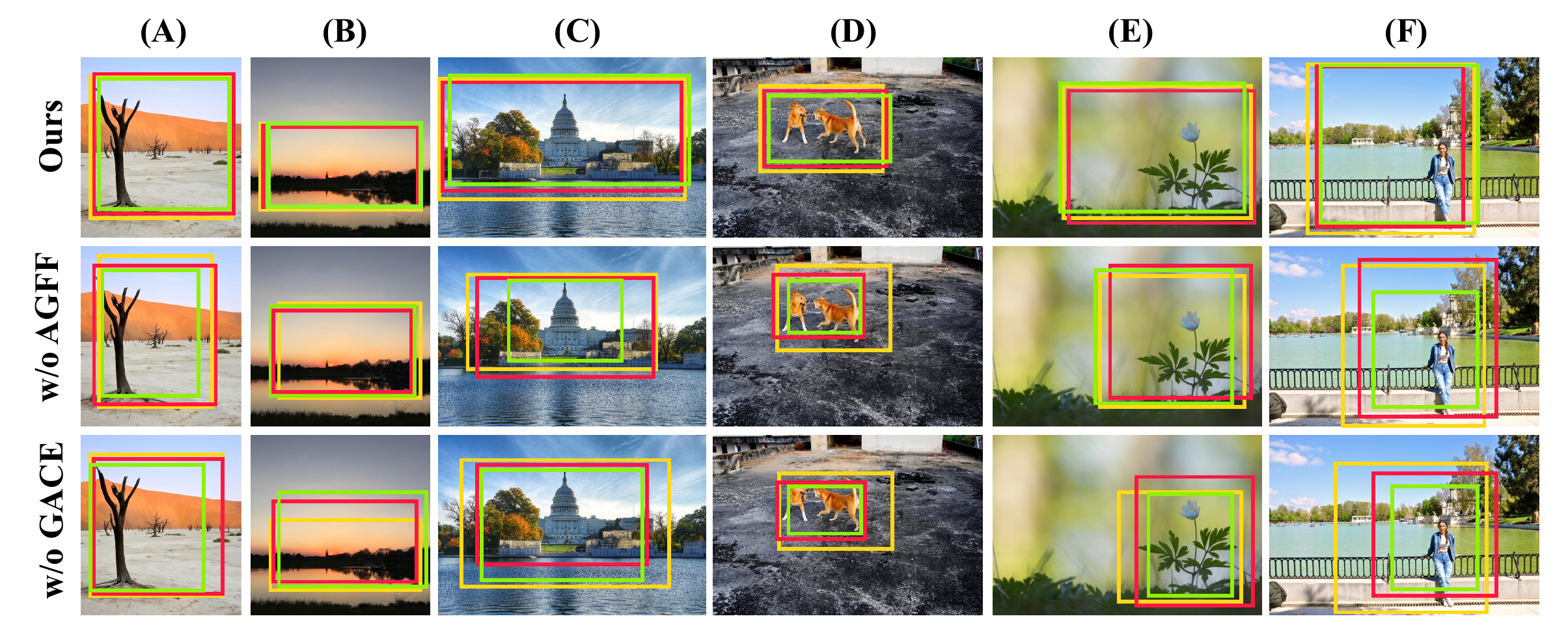}
    \caption{Top three crop planting areas predicted under different model conditions in the ablation experiment. The red box represents the best planting area, the yellow box the second best, and the green box represents the third best.}
    \label{fig:temporal_stability}
\end{figure}

This section presents qualitative ablation experiments to further and intuitively verify the role of each module in the generated results. We analyzed two key dimensions: coarse attention distribution in different regions and perception ability of boundary regions. To illustrate the impact of each component, we compared the results obtained after progressively removing the AGFF and GACE modules with those generated by the complete model. The results are shown in Figs.~\ref{fig:ablation}--~\ref{fig:temporal_stability}. For demonstration purposes, we randomly selected images from different categories.

\textbf{Coarse attention distribution.} The cropped image fails to reflect the importance of different local regions when the AGFF module is removed. Fig. \ref{fig:ablation} shows the optimal cropping results after removing different modules. The yellow box represents the ground-truth annotation, while the red box indicates the prediction. We can clearly observe that for images with obvious objects in columns (A), (D), (E), and (F), the optimal cropping results do not fully preserve the object information. For example, the person's feet are cropped out in (F). A similar pattern appears in images without obvious objects in columns (B) and (C). For example, the lower right corner of the main building is missing in (C), and the plant background in the upper left corner is not fully preserved, failing to accurately capture the important content of the image. This experiment shows that the AGFF module enables accurate and effective cropping.

\textbf{Perception in boundary regions.} If the GACE module is removed, the prediction order of multiple cropping boxes may fluctuate significantly, even when the content within the cropping boxes is roughly the same. Fig.~\ref{fig:temporal_stability} shows the top three prediction results after removing different modules. The first row shows the prediction results produced by our method. We can clearly observe that the images in columns (A), (D), (E), and (F) contain prominent objects. Although some object information remains after removing different modules, the top three predictions from our method are closer to the ground-truth annotations. For example, for image (F), the candidate box predicted without the GACE module is significantly different from the annotation. In contrast, our method produces candidate boxes whose proportions and areas closely match the annotated values. A similar pattern is observed for images without obvious objects in columns (B) and (C). For example, for image (B), the yellow candidate box predicted after removing the GACE module is significantly inconsistent with the other two results. By comparison, the top three candidate boxes predicted by our method align closely with the annotated content.

In summary, both the AGFF and GACE modules enhance quantitative metrics and qualitative evaluations, thus improving cropping accuracy and ranking stability. These advantages further validate the effectiveness of our method.

\subsection{Hyper-parameter Analysis}

\begin{figure}
    \centering
    \includegraphics[width=0.6\textwidth]{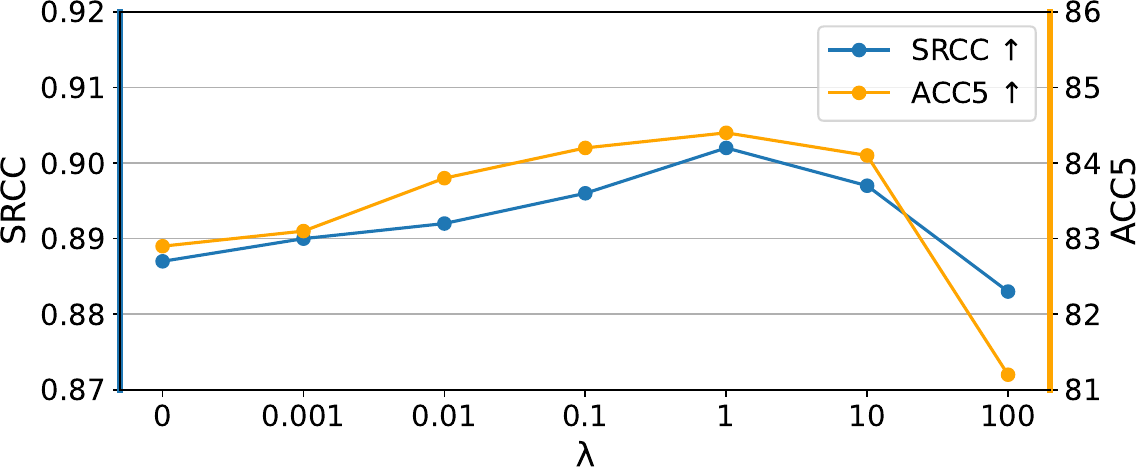} 
    \caption{Our method investigated the impact of different $\lambda$ values on SRCC and ACC5 on the GAICD dataset.} 
    \label{fig:para}
\end{figure}

\begin{figure}
    \centering
    \includegraphics[width=0.6\textwidth]{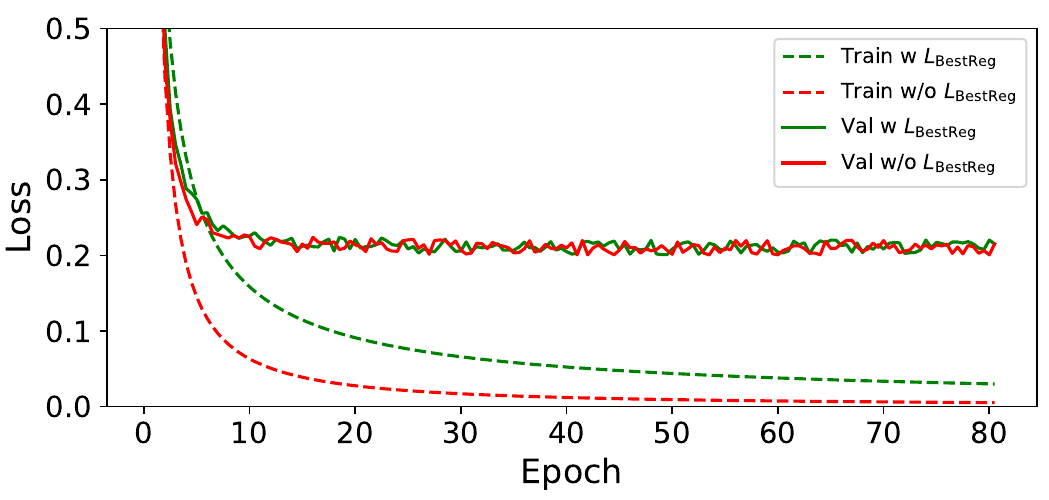} 
    \caption{Training and validation loss with and without $L_{\text{BestReg}}$} 
    \label{fig:para1}
\end{figure}

The optimal loss function, $\mathcal{L}_{BestReg}$, is proposed to represent the spatial coverage consistency between the highest-scoring predicted candidate box and the manually labeled optimal box. To verify the effectiveness of this loss function, we conducted an experimental analysis, and the results are shown in Figs. \ref{fig:para}-\ref{fig:para1}.


In Eq. (13), we introduce a parameter $\lambda$ to calculate $\mathcal{L}_{opt}$, adjusting for the loss associated with the optimal box score. Fig. \ref{fig:para} shows SRCC and ACC5 for different $\lambda$ values. When $\lambda=1 $, our algorithm achieves joint optimization of scoring accuracy, ranking consistency, and optimal result matching through the total loss function. 

Figure \ref{fig:para1} shows the learning curves for our model trained with and without $\mathcal{L}_{BestReg}$, using 100 randomly selected images from the training set for validation. Without $\mathcal{L}_{BestReg}$, the training loss  decreases rapidly, but a significant gap remains compared to the validation loss. In contrast, incorporating $\mathcal{L}_{BestReg}$ results in a smaller validation loss and a reduced difference between training and validation losses, thus improving the model’s generalization capability.

\section{Conclusion}
\label{dc} 

This paper proposes a general aesthetic image cropping framework, namely Global Attention-Fused Image Cropping (GAFIC), which emphasizes precise attention distribution and enhanced perception of boundary regions. First, GAFIC utilizes the Attention-Guided Feature Fusion (AGFF) module to hierarchically weight image regions according to their importance. These weights are combined with the image’s feature vectors to generate a novel feature representation. By refining weight differences between regions, the model achieves a more precise attention distribution for fine-grained regions. Second, GAFIC uses the Global-Aligned Crop Evaluator (GACE) module to fuse internal and external features of each candidate bounding box with global image features for aesthetic score prediction. This fusion enables the evaluation process to better capture global context, thereby enhancing perception near boundaries. Finally, we introduce a multi-dimensional loss function that jointly optimizes the regression score of the optimal cropping region, the overall regression score of all cropping regions, and the ranking score. This ensures that training focuses on achieving the best possible results. In our experiments, we selected highly complex images that cover diverse scenes, such as people, animals, architecture, and natural landscapes, as target materials. Through extensive quantitative and qualitative comparisons, we demonstrate the superiority of  GAFIC. Compared to existing methods, our method more accurately and consistently identifies aesthetically pleasing regions in images.

GAFIC follows the image-cropping paradigm and is not designed for retargeting tasks that require content synthesis, content-aware filling, or arbitrary aspect-ratio transformation. The AGFF heat map is constructed from candidate crop boxes, so its quality can be affected by the candidate-generation strategy and may need further validation in domains with different spatial layouts. The current implementation uses VGG16 as the backbone, and stronger encoders may further improve feature representation, which will be explored in future backbone ablations. Finally, the present evaluation protocol does not stratify test images by annotation agreement or crop-quality ambiguity; therefore, the reported average metrics should be interpreted as overall benchmark performance rather than as a complete analysis of highly ambiguous cropping scenes.

\section{Acknowledgments}
This work was supported by the Science and Technology
Development Fund, Macao SAR, under the Basic Research Program
(0006/2024/RIA1); the Science and Technology Development Fund--Ministry of Science and Technology under the National Key R\&D Program of China (0007/2025/AMJ and 2025YFE0202900); the Shenzhen Science Fund for Excellent Young Scholars (RCYX20221008093036022); the Special Support Plan for Outstanding Young Talents of Guangdong Province (2023TQ07L745); and the Youth Innovation Promotion Association of the Chinese Academy of Sciences (2021358).
\bibliographystyle{elsarticle-num}
\bibliography{references}

\end{document}